# An Explainable AI Model for Predicting the Recurrence of Differentiated Thyroid Cancer

Mohammad Al-Sayed Ahmad
*Department of Medical Engineeringg*
*Al-Ahliyya Amman University*
Ammaan, Jordan
20191169@ammanu.edu.jo

Jude Haddad
*Department of Medical Engineeringg*
*Al-Ahliyya Amman University*
Ammaan, Jordan
202327152@ammanu.edu.jo

*Abstract*—Thyroid carcinoma, a significant yet often controllable cancer, has seen a rise in cases, largely due to advancements in diagnostic methods. Differentiated thyroid cancer (DTC), which includes papillary and follicular varieties, is typically associated with a positive prognosis in academic circles. Nevertheless, there are still some individuals who may experience a recurrence. This study employs machine learning, particularly deep learning models, to predict the recurrence of DTC, with the goal of improving patient care through personalized treatment approaches. By analysing a dataset containing clinicopathological features of patients, the model achieved remarkable accuracy rates of 98% during training and 96% during testing. To improve the model's interpretability, we used techniques like LIME and Morris Sensitivity Analysis. These methods gave us valuable insights into how the model makes decisions. The results suggest that combining deep learning models with interpretability techniques can be extremely useful in quickly identifying the recurrence of thyroid cancer in patients. This can help in making informed therapeutic choices and customizing treatment approaches for individual patients.

## I. Introduction

Thyroid carcinoma constitutes about 1% of all newly diagnosed malignancies, making up around 0.5% of malignancies in males and 1.5% in women. Differentiated thyroid carcinomas (DTC) account for 94% of cases, mainly papillary (PTC) or follicular (FTC) carcinomas [1]. The global prevalence has risen to 25.18 per 100,000, and 29.45 per 100,000 in the MENA region [2]. Despite low mortality, DTC recurrence remains a concern, necessitating predictive models for personalized treatment and follow-up strategies [3]. Differentiated thyroid carcinoma generally has a high long-term survival rate, but some patients with PTC and a larger minority with FTC may die from the disease. Prognostic models consider factors like age, gender, tumour characteristics, extrathyroidal extension, clinical stage, and distant metastases to identify high-risk patients [4-6]. Papillary thyroid carcinoma, the most common subtype of DTC, has increased by 3% annually in the U.S. Treatment typically involves thyroidectomy and Radioactive Iodine (RAI). Although most DTCs are low-risk, about 30% may recur, necessitating further treatment, such as RAI, surgery, or chemotherapy, which can affect patient quality of life [7].

DTC diagnosis and treatment rely on histological categories and genetic mutations. Conventional risk stratification offers static risk assessments, while modern models provide dynamic, continuously updated projections, improving long-term prediction accuracy for patients [8].

Machine learning (ML), a branch of artificial intelligence (AI), analyses data to identify patterns, offering insights into complex health scenarios like disease risk factors and prognosis. Unlike traditional programming, ML uses learning algorithms and probabilistic models for data-driven predictions. Deep learning (DL), an advanced ML form, leverages multi-layered neural networks, benefiting from increased computing power and larger datasets, enhancing its predictive capabilities [9].

Deep learning has revolutionized cancer prediction, significantly improving diagnostic accuracy and efficiency. Advanced algorithms now enable predictive models to analyse complex medical data with minimal human intervention. For instance, deep learning has enhanced mammographic breast density classification, crucial for breast cancer diagnosis, and improved early detection and treatment outcomes. Additionally, deep learning techniques applied to lung cancer whole-slide imaging and glioblastoma survival predictions demonstrate its potential in guiding treatment strategies and advancing personalized oncology [10].

As black-box ML models gain use in critical predictions, transparency has become crucial. Without explainable decisions, these models risk losing legitimacy and trust, particularly in precision medicine, autonomous vehicles, security, and finance, where interpretability is essential [11].

Explainable AI (XAI) addresses the trade-off between model performance and transparency by creating interpretable models without sacrificing learning performance. XAI enhances understanding, trust, and management of AI systems, ensuring impartial decision-making, detecting biases, improving robustness, and establishing true causality in model reasoning. Interpretability provides insights into mechanisms, visualizes discrimination rules, and identifies potential perturbations [11].

XAI offers ML techniques that balance high performance with explainability, drawing on social sciences and psychology to create transparent, trustworthy AI systems. This approach enhances AI adoption and effectiveness in critical domains [11].

Li et al. (2021) explore AI's potential to enhance personalized thyroid cancer treatment by improving diagnosis, risk stratification, and therapy through analysis of morphological, textural, and molecular features. AI can outperform humans in evaluating thyroid nodules using ultrasonography, cell smears, and tissue slices, feeding this data into AI classifiers for better accuracy. However, challenges include the need for large, high-quality datasets, lack of explainability, financial burdens, and validation due to cancer variability. Despite these limitations, AI holds promise for addressing key issues in personalized thyroid cancer therapy by enabling more accurate and tailored treatments [12].

Olatunji et al. (2021) evaluated machine learning tools for early thyroid cancer detection, finding that the random forest (RF) method achieved the highest accuracy of 90.91% using seven features. The study compared RF, artificial neural networks (ANN), support vector machines (SVM), and Naive Bayes (NB), using data preparation techniques like feature selection and grid search. Despite limitations, such as not incorporating image-based features, RF proved the most effective for early detection [13].

Habchi et al. (2023) provide a comprehensive analysis of AI applications in thyroid cancer, exploring supervised (deep learning, neural networks, traditional classification), unsupervised (clustering, dimensionality reduction), and ensemble methods (bagging, boosting). The study reviews datasets, feature selection, and performance criteria, noting challenges like scarce well-curated datasets, limited annotated data, and the need for balance between accuracy and processing time. Despite AI's potential to improve diagnostic accuracy, issues such as dataset imbalance and computational complexity persist, emphasizing the need for better datasets, improved imaging techniques, and multidisciplinary collaboration for effective AI integration in healthcare [14].

Verburg and Reiners (2019) explore using AI and deep learning to analyse thyroid ultrasound images for cancer detection. They propose a multinational study framework but note limitations, including the lack of integration with non-imaging clinical data and detailed methods. The study emphasizes using extensive training datasets, incorporating advanced ultrasound techniques like colour Doppler and 3D imaging, and improving methodological rigor. Future research should include well-defined cases, controls, and biomaterial samples to enhance AI models' accuracy for thyroid cancer detection [15].

Lastly, the publication by Borzooei et al. (2023) provides a detailed account of developing and evaluating machine learning algorithms for predicting recurrence probability in patients with well-differentiated thyroid carcinoma. The study analysed a retrospective cohort of 383 patients over 15 years, collecting 13 clinicopathologic characteristics. Various machine learning models (K-nearest neighbours, support vector machines, tree-based models, and neural networks) were evaluated using three datasets. Internal validation with 100 patients measured sensitivity, specificity, and accuracy. Constraints included the lack of external validation, potential overfitting, and biases due to limited clinical subtypes and early-stage diagnoses. The study highlights the potential of machine learning in predicting cancer recurrence and customizing treatment and follow-up intervals for patients with well-differentiated thyroid cancer [3].

## II. METHODOLOGY

### A. Dataset

The dataset, available on the UCI Machine Learning Repository, addresses the recurrence of differentiated thyroid cancer. It includes 383 instances and 16 diverse features (real, categorical, integer) relevant to thyroid cancer recurrence, such as age, gender, smoking history, and tumour characteristics. Collected over 15 years with a minimum 10-year follow-up, this tabular dataset offers a comprehensive view of clinical and pathological factors, facilitating the development of predictive models for assessing recurrence risk [16].

### B. Data Pre-processing

The dataset comes as a CSV file with various variables relevant to thyroid cancer, including demographic data, medical history, and clinical assessments. The goal variable, "Recurred," indicated cancer recurrence and served as the target for predictive modelling. Data pre-processing involved converting categorical variables into numerical values through label encoding, which made the data suitable for machine learning. The dataset was split into features (X) and the target variable (y), with "Recurred" as the target. The data was further divided into 80% training and 20% testing subsets. Features were standardized using a Standard Scaler to ensure each feature had a mean of zero and a standard deviation of one, which enhances neural network performance and training efficiency.

### C. Model Building, Training, and Evaluation

The model was developed using TensorFlow's Keras API with a deep neural network architecture designed to capture complex data patterns. It included an input layer, three Dense layers with 128, 64, and 32 units, each followed by a Dropout layer at 0.5 to mitigate overfitting. ReLU activation functions were employed in the Dense layers to introduce non-linearity, and the output layer had a single Dense unit with a sigmoid activation function for binary classification of cancer recurrence. The Adam optimizer with a learning rate of 0.001, binary cross-entropy loss, and accuracy as the performance metric were used. Training lasted 100 epochs with a batch size of 32, and validation data (20% of the test set) helped monitor performance and prevent overfitting.

Following training, the model's performance was assessed using various metrics. Accuracy measured the proportion of correctly classified instances, while sensitivity (recall) indicated the percentage of actual positives detected, and specificity assessed the true negative rate. Positive predictive value (PPV) and negative predictive value (NPV) evaluated the accuracy of positive and negative predictions, respectively. These metrics were applied to both training and testing datasets to assess performance on new data. A confusion matrix was also generated to visually display true positives, true negatives, false positives, and false negatives, providing further insight into the model's effectiveness

*D. Model Interpretability*

In medical applications, understanding the decision-making process of neural networks is crucial for building confidence and ensuring interpretability. To address this, two interpretability methods were employed: Local Interpretable Model-Agnostic Explanations (LIME) and Morris Sensitivity Analysis.

LIME was used to generate localized explanations for the model's predictions by creating a simplified model that approximates the complex model for specific predictions. This approach identifies the features most influential in a particular prediction. A LIME explainer was developed using the training data, tailored for classification tasks, and included the ability to discretize continuous features. A custom `predict_proba` function was created to compute class probabilities for LIME. An example from the test set was explained, revealing significant insights into the model's decision-making process.

Morris Sensitivity Analysis was performed to complement LIME with global sensitivity analysis. This technique identified influential features across the entire dataset, providing a comprehensive view of each feature's impact on model predictions. Variables were clearly defined, and a wrapper function was created for predictions within the sensitivity analysis framework. Sensitivity indices ($\mu^*$, $\sigma$) were calculated, offering insights into the mean and variability of each feature's effect, thus enhancing understanding of the model's behaviour.

### III. P RESULTS AND DISCUSSION

*A. Results of the model*

The predictive model developed for detecting thyroid cancer recurrence demonstrated exceptional performance, highlighting its potential for clinical use. Trained and validated on a comprehensive dataset with diverse clinicopathologic characteristics, the model achieved impressive accuracy rates. It reached 98% accuracy during training depicted in "Fig. 1" and "Fig. 2", reflecting its ability to learn effectively from the data. Its testing accuracy was 96%, indicating strong generalization to new data.

The confusion matrix (shown in "Fig. 3") from the test set provided further insights into the model's performance. It correctly identified 16 out of 19 true recurrences (true positives) and missed 3 (false negatives). It perfectly detected all 58 non-recurrences (true negatives) with no false positives, underscoring its strong specificity and ability to minimize false alarms—critical in clinical settings where false positives can lead to unnecessary procedures and patient anxiety.

Performance metrics (listed in Table I) from the training phase revealed a sensitivity of 97.75%, showing the model's proficiency in detecting recurrences, and a specificity of 100%, meaning it accurately identified all non-recurrences. The positive predictive value (PPV) was 100%, indicating that all predicted recurrences were accurate. The negative predictive value (NPV) was 99.09%, demonstrating high accuracy in predicting non-recurrences.

In the test set, sensitivity slightly decreased to 84.21%, but specificity remained perfect at 100%. The PPV continued to be 100%, confirming that all predicted recurrences were genuine, while the NPV was 95.08%, reflecting high accuracy for non-recurrences.

The results showcase the model's strengths and areas for improvement. While the model excelled in sensitivity during training, the drop in test sensitivity suggests challenges in applying its capabilities to new data, possibly due to differences between the training and test sets. However, its consistent specificity and high predictive values in both phases were commendable.

Overall, the model's high accuracy, specificity, and predictive values make it a valuable tool for predicting thyroid cancer recurrence. It can assist physicians in identifying at-risk patients with confidence, allowing for timely interventions. Future work should focus on enhancing sensitivity, potentially through more diverse training data or advanced techniques to improve generalization.

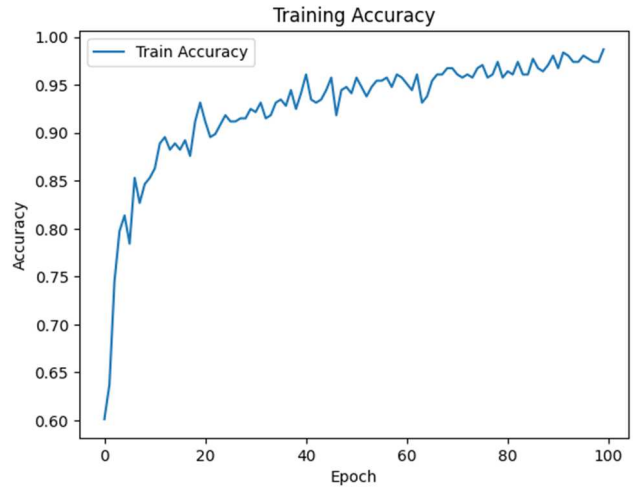

Fig. 1. Training accuracy

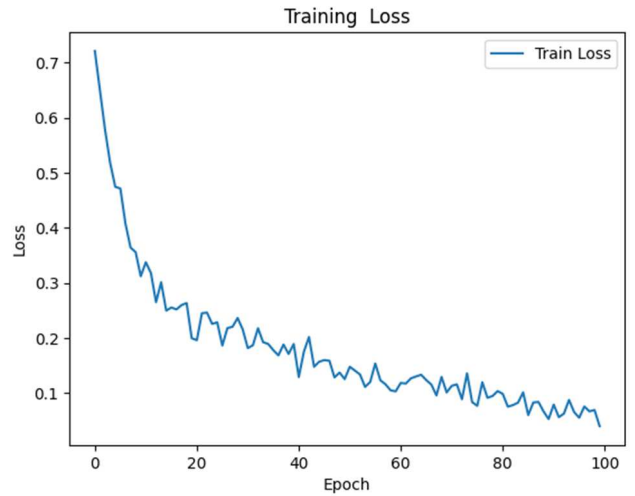

Fig. 2. Training loss

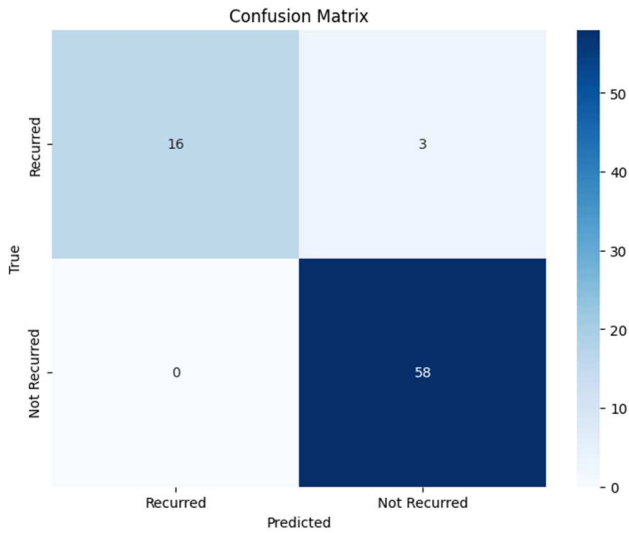

Fig. 3. Test set confusion matrix

TABLE I. TRAINING AND TESTING METRICS

|  | Train | Test |
|---|---|---|
| **Sensitivity** | 0.978 | 0.842 |
| **Specificity** | 1.0 | 1.0 |
| **PPV** | 1.0 | 1.0 |
| **NPV** | 0.991 | 0.951 |

*B. LIME Results*

LIME provided clear insights into individual thyroid cancer recurrence predictions, examining cases with 100% certainty of non-recurrence, 100% certainty of recurrence, and 78% non-recurrence versus 22% recurrence likelihood.

In the first case, the model predicted with 100% certainty that the patient would not relapse. LIME analysis highlighted key factors: a high Thyroid Function value (3.21) and a low Pathology score (-2.79). High Thyroid Function indicated improved health, and a low Pathology score suggested fewer abnormalities, both supporting the no-recurrence prediction. This analysis confirmed the model's confidence in the low-risk profile and no expected recurrence. "Fig.4" clarifies how each factor influenced the decision.

In the second case shown in "Fig. 5", the model predicted a 100% chance of recurrence. LIME analysis identified high M (4.75) and T (2.08) values, indicating advanced metastasis and tumour size, and a negative Risk value (-2.40) as key factors in this prediction.

In the third example, the model predicted a 78% chance of no recurrence and a 22% chance of recurrence "Fig. 6". LIME analysis showed that while most characteristics suggested non-recurrence, high risk variables like Adenopathy (1.73) and Physical Examination (1.07) introduced uncertainty, demonstrating the model's ability to handle complex data and emphasize the need for careful monitoring.

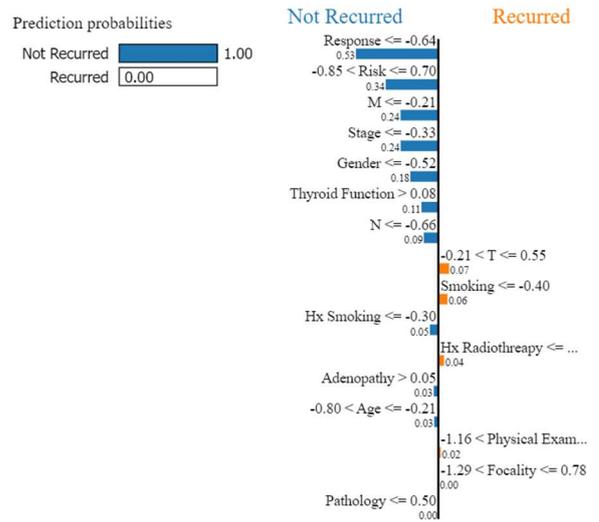

Fig. 4. LIME analysis of "Not Recurred" case

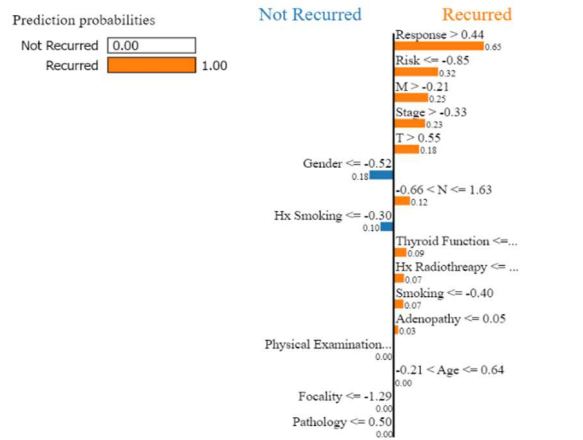

Fig. 5. LIME analysis of "Recurred" case

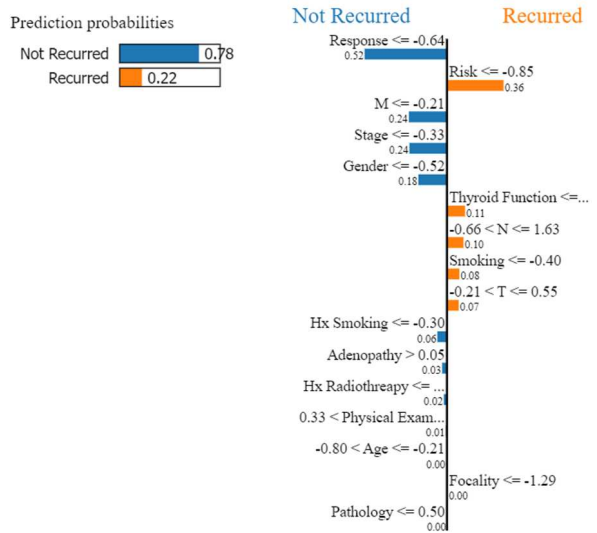

Fig. 6. LIME analysis of "Undecided" case

## C. Morris Sensitivity Analysis Results

The Morris Sensitivity Analysis provided a comprehensive understanding of the features influencing the thyroid cancer recurrence prediction model. Two key metrics were used: μ* (mu_star), which represents the average impact of each feature on the model's output, and σ (sigma), which indicates the variability of these impacts due to interactions with other features.

Among the variables, Response showed the highest μ*, indicating it had the greatest influence on predictions. Stage also had a significant μ*, highlighting its crucial role in decision-making. Moderate σ values for these features suggest they interact with other variables, adding complexity to the model.

Tumour Size (T) and Adenopathy had moderate μ* values and high σ values, reflecting their substantial but variable impact depending on other factors. The elevated σ values point to complex interactions affecting their influence.

Physical Examination and Risk had moderate μ* values, with Physical Examination showing a moderate σ and Risk a low σ. These features contributed consistently to the model's predictions with less variability, indicating their supportive roles in enhancing prediction accuracy.

M (Metastasis) and N (Nodal Involvement) showed lower μ* and σ values, implying a modest but consistent impact on predictions. Despite their lower ranking, they were significant in specific contexts.

Pathology, Focality, and Thyroid Function had low μ* and σ values, suggesting minimal impact on predictions compared to other features illustrated in "Fig. 7".

Overall, the analysis underscored the importance of Response and Stage as primary predictors, while Tumour Size and Adenopathy demonstrated complex interactions. Risk and Physical Examination were reliable and stable, contributing consistently to the model. The Morris Sensitivity Analysis clarified the roles of different features, helping translate model predictions into practical clinical insights.

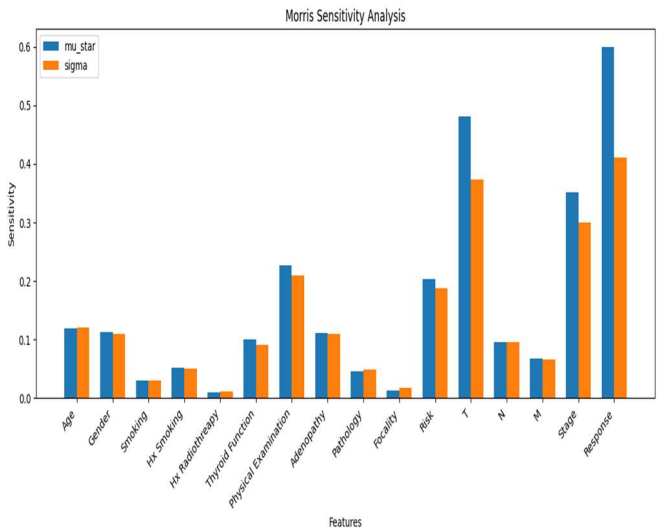

Fig. 7. Morris Sensitivity Analysis

## IV. CONCLUSION

Utilizing deep learning models to forecast the reappearance of differentiated thyroid cancer signifies a substantial advancement in tailored medical therapy. The model used in this study exhibited exceptional accuracy, highlighting its potential significance in clinical practice for improving patient outcomes. By including interpretability techniques like LIME and Morris Sensitivity Analysis, the model's dependability is enhanced as it makes its decision-making process clearer. This transparency is crucial for winning the confidence of both physicians and patients. Nevertheless, the study highlights the significance of additional validation in other patient groups and the incorporation of more comprehensive, high-quality datasets to strengthen the model's reliability and applicability. Subsequent investigations should prioritize surmounting these obstacles, so promoting the broader use of AI-powered instruments in the realm of cancer.